# Towards Reversible De-Identification in Video Sequences Using 3D Avatars and Steganography

Martin Blažević*, Karla Brkić*, Tomislav Hrkać*
*University of Zagreb, Faculty of Electrical Engineering and Computing, Croatia
Email: `martin.blazevic@fer.hr`, `karla.brkic@fer.hr`, `tomislav.hrkac@fer.hr`

*Abstract*—We propose a de-identification pipeline that protects the privacy of humans in video sequences by replacing them with rendered 3D human models, hence concealing their identity while retaining the naturalness of the scene. The original images of humans are steganographically encoded in the carrier image, i.e. the image containing the original scene and the rendered 3D human models. We qualitatively explore the feasibility of our approach, utilizing the Kinect sensor and its libraries to detect and localize human joints. A 3D avatar is rendered into the scene using the obtained joint positions, and the original human image is steganographically encoded in the new scene. Our qualitative evaluation shows reasonably good results that merit further exploration.

## I. Introduction

Protecting the privacy of law-abiding individuals in surveillance footage is becoming increasingly important, given the widespread use of video surveillance cameras in public places such as streets, transit areas, shopping centers, banks, etc. Unrestricted access to such footage without privacy protection could enable identification and tracking of individuals in real time. In practice, privacy protection of persons in image and video data can be achieved using de-identification techniques. De-identification is a process of reversibly removing or obfuscating identifying personal biometric features, such as skin, hair and eye color, body posture, clothing, birthmarks, tattoos etc.

We envision a de-identification pipeline that enables reversible concealment of persons' identities in video sequences. Our pipeline aims to de-identify all hard biometric features, while taking into account most soft biometric features (e.g. tattoos, birth marks, clothing). The steps in our pipeline are as follows: (i) detecting humans in video sequences, (ii) segmentation of the detections, (iii) human pose estimation (joint modeling), (iv) rendering naturally-looking 3D models of humans on top of human detections, and (v) steganographically encoding the de-identified information in the video, as illustrated in Figs. 1 and 2. Building this pipeline is our long term goal, and in this work we focus on qualitatively exploring the last two stages of the pipeline, i.e. (iv) rendering naturally-looking 3D models of humans on top of human detections and (v) steganographically encoding the de-identified information in the video. Previous stages of the pipeline have been explored in our work [1], [2].

## II. Related work

Each stage of the proposed detection pipeline is a broad topic of research in computer vision, so we limit ourselves to briefly reviewing methods that could be applicable in individual stages. Note that in a practical de-identification system one should expect the boundaries between the first three stages to become blurred: detecting, segmenting and estimating the pose of humans can often become an iterative process where the stages collaborate to find the most likely hypothesis.

Modern methods for pedestrian detection include the HOG detector [3] (histograms of oriented gradients), deformable part models [4], integral channel features [5], [6], convolutional neural networks [7], [8], [9], the Viola-Jones detector [10] and its problem-specific extensions [11], etc. Using background subtraction [12], [13] or recently introduced object proposals [14] for person detection might also be considered if the problem scenario is sufficiently constrained. Jointly employing several methods could be a promising strategy, as recent comparative studies [5], [15], [6] have shown that pedestrian detection in non-ideal conditions (moving camera, small-scale pedestrians, occlusions) is a hard problem that benefits from combining multiple features.

Having a detection bounding box, fine segmentation-based localization of pedestrians can be achieved by e.g. one of the variants of the GrabCut segmentation algorithm [16]. Furthermore, detection, tracking and segmentation can be done in a bootstrapping loop, as e.g. in the work of Hernandez-Vela et al. [17]. They propose automating GrabCut by combining HOG-based human full-body detection and face and skin color detection with tracking and segmentation models. In similar spirit, Poullot and Satoh [18] propose an algorithm called VabCut, useful for video foreground object segmentation when the camera is not stationary, where object locations are inferred using motion.

Human pose estimation can be combined with detection in a part-based model framework, such as in the work of Andriluka et al. [19] who propose a general part-based approach based on pictorial structures. Similarly, Yang and Ramanan [20] use a flexible mixture model that encodes both co-occurences and spatial relations between human body parts. Toshev and Szegedy [21] propose estimating human pose using deep learning, formulating it as a regression problem towards body joints. In general, human pose estimation is difficult, as each joint has many degrees of freedom, and the appearance of body parts can vary considerably due to clothing, body shape, viewing angle etc. [20]. The problem can be somewhat simplified by introducing another layer of data: the depth map of the scene, and using a dedicated sensor such as Microsoft Kinect. Shotton et al. [22] introduce the Kinect's pose estimation algorithm based on randomized decision forests, body part recognition and depth image features. The algorithm achieves acceptable mAP rates and works very fast on dedicated hardware.

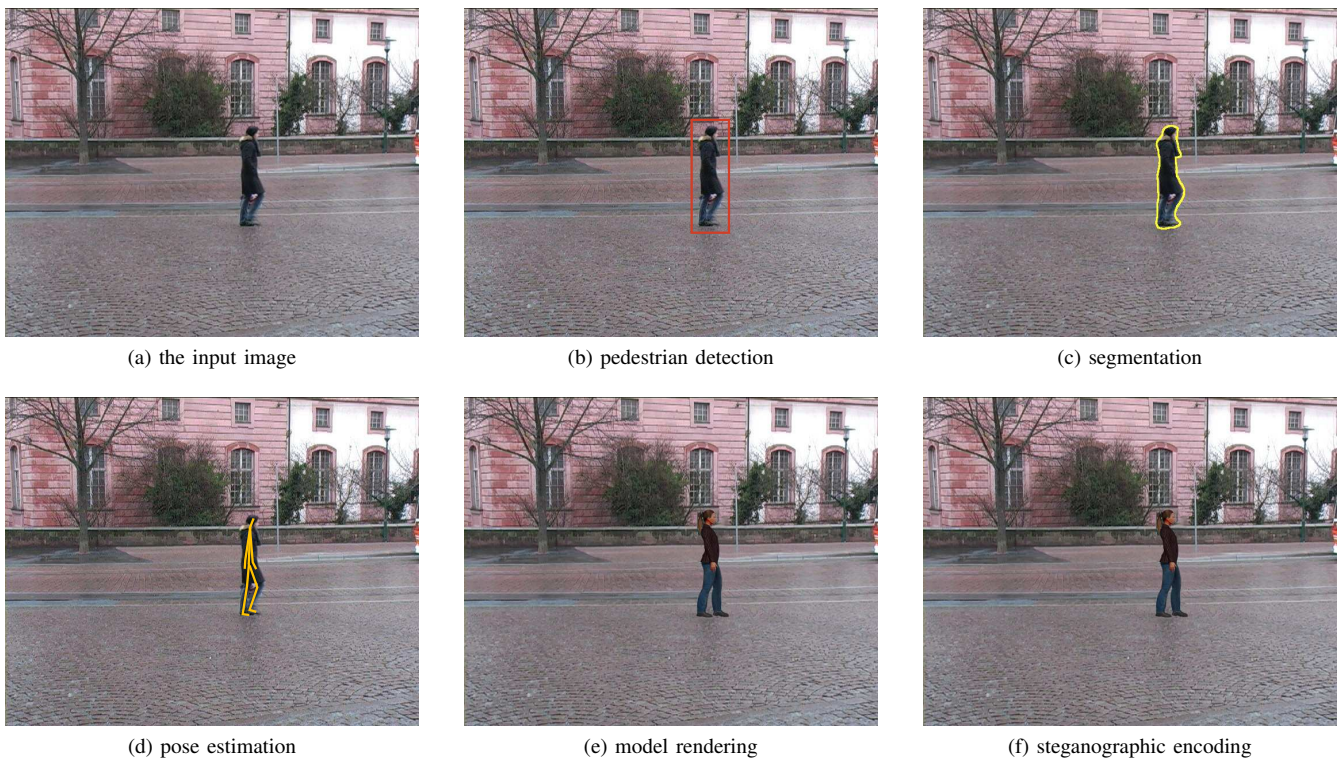

Fig. 1: The visualisation of our de-identification pipeline.

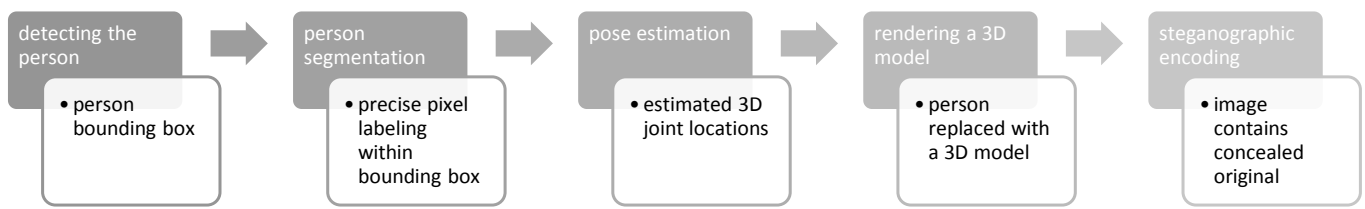

Fig. 2: The steps in our envisioned de-identification pipeline.

With localized joints and precise positions of human body parts, it is possible to replace portions of or the entire rendered human with an artificial realistically-looking model. The majority of related work in this area focuses on face replacement. For instance, Blanz et al. [23] propose a face exchange algorithm that enables fitting morphable face models to an image, effectively replacing the face in the image with the rendered model, taking into account pose and illumination parameters. Dale et al. [24] propose a system for face replacement in videos, enabling replacing the face and its motions due to speech and facial expressions with another face. Samaržija and Ribarić [25] propose using active appearance models for face detection and de-identification.

Finally, the information concealed by rendering the 3D model can be stored hidden in the new image using steganography, a technique used for secretive communication by hiding information in digital media. Steganography is also referred to as the art and science of communicating which hides the existence of the communication [26]. Steganographic encoding of images aims to exploit the image format in order to store the secret information so that the changes to the image are imperceptible. Various algorithms exist [27], and in this work we focus on LSB insertion [28] that encodes the secret information by modifying the least significant bits of the carrier image.

### III. THE ENVISIONED DE-IDENTIFICATION PIPELINE

Our long-term goal is building a de-identification pipeline that enables privacy protection of persons in video sequences by rendering a naturally-looking 3D human model in place of each detected person. Simultaneously, the image of the per-son is steganographically encoded in the replacement image. Hence, the naturalness of the scene is preserved, while the private information is stored in a secure manner.

In this work, we focus on exploring the last two stages of the pipeline, namely replacing the person in the video with a 3D model and encoding the replaced image in the new image using steganography. In order to be able to render a 3D model at the correct place, we must first detect and localize the person and estimate her pose, i.e. find the 3D positions of joints. As mentioned previously, this is a hard problem [20], [15], therefore, in this work we investigate the general feasibility of our approach by choosing a close to best case scenario. We ask ourselves how well the 3D model rendering and steganographic encoding would perform if the person is reliably detected and her joint positions are localized. To ensure this scenario, we use dedicated hardware, namely the Microsoft Kinect sensor. Using Kinect SDK, we are able to obtain a reasonably reliable pose estimation of a human standing close to the sensor. We now describe the process of pose estimation, 3D model rendering and steganographic encoding in more detail.

*A. Recovering the 3D skeleton from Kinect data*

Microsoft Kinect is a multifunctional physical device which enables the detection of human voice, movement and face gestures. It combines an RGB camera, 3D depth sensor cameras, a microphone array and a software pipeline for processing color, depth, and skeleton data [29]. The RGB camera enables capturing a color image. Depth sensor cameras consist of an infrared (IR) emitter and an IR depth sensor, which enable capturing a depth image. The distances between the observed object and the sensor itself represent the depth information. This type of information is used for detecting and precise tracking of human body or skeleton.

The Kinect for Windows Software Development Kit (SDK) is a set of drivers, Application Programming Interfaces (APIs), device interfaces, and tools for developing Kinect-enabled applications [30]. The Kinect sensor and its software libraries use the Natural User Interface (NUI) libraries to interact with custom applications. NUI represents the core of the Kinect for Windows API. It supports fundamental image and device management features. It also makes sensor data, such as color and depth image data, audio data, as well as skeleton data, which Kinect sensor generates in real time by processing the depth data, accessible from inside user developed applications. There are a few configuration prerequisites for applications that use or display data from the skeleton stream, such as the one provided in this paper. The used streams should be identified, initialized and opened. Skeletal tracking must also be enabled. Buffers for holding sensor data should be allocated and the new data retrieved for each of used streams every new frame.

Kinect is able to detect and recognize up to six users standing in front of the sensor, or rather in the field of view (FOV) of the camera, and accurately track, extract and provide detailed information on up to two of them. For a user to be tracked, no specific pose, gesture or calibration action needs to be performed.

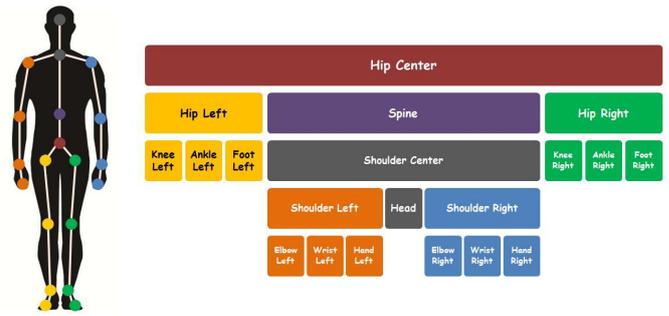

Fig. 3: The hierarchy of the joints in Kinect skeletal tracking.

Hierarchy of joints defined by the skeletal tracking system is shown in Fig. 3. Unlike joints, bones are not explicitly defined in the API. Each bone is uniquely specified by the parent and the child joints that connect the bone. Bones rotation and movement follows the movement and orientation of its child joint. Joint *HipCenter* is the root joint and parent to all other, children joints. Other joints can have multiple parents. For example, joint *ShoulderCenter* has *HipCenter* as one of its parents and *Spine* joint as other.

*B. Rendering the 3D model*

In this section we explain how the movement of the person in front of the camera can be transferred to the movement of the virtual 3D model in the virtual scene by using Kinect. The model itself is created through a process of forming the shape of an object. It can be made in software which enables modeling, simulation, animation and rendering, such as Maya, Blender, MakeHuman and others. A 3D model is essentially a mathematical representation of a 3D object. It is a set of points in 3D space interconnected by lines, curves, surfaces, etc.

The virtual model used in this paper represents a virtual character or avatar, comprised of a skeleton and a mesh. The skeleton has one bone for each movable part of the model and it is used to animate the avatar [31]. The skeleton does not have a graphical representation, rather each bone is linked to a defined set of vertices. A collection of interconnected vertices defines the character's visible surface or skin of the model – mesh, which appears in the rendered frame. The process of building the character's armature or skeleton and setting up its segments to ease the animation process is called *rigging*. The process of attaching the defined bones to a mesh is called *skinning*. Rigging and skinning are prerequisites for model to be used in *skeletal animation* – a technique of animating the defined model used in this paper.

Nowadays, skeletal animation is widely used because it facilitates the often highly complex process of animating human characters by providing a simplified user interface. The implementation used in this paper is based on Kinect SDK, and it uses an extended version of Content Pipeline called *Skinned Model Pipeline* or *Skinned Model Processor*. This pipeline enables 3D models, textures, animations, effects, fonts and similar to be imported to the project and processed. Once loaded, the 3D model can be rendered and animated.

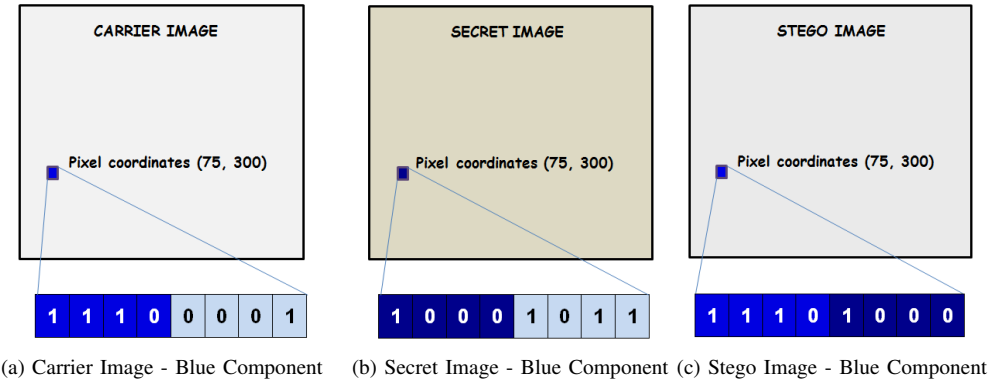

Fig. 4: The process of creating a stego image.

Our solution enables the model in the scene to mimic the movements and gestures of people that are in the FOV of the Kinect camera. By using the Video data API (Data sources and Streams) the application can locate and retrieve information for twenty joints of the tracked user and compute real-time tracking information. The avatar is animated using the skeleton stream. The processing algorithm handles all of tracked skeletons and maps the movement of tracking to a movements of 3D model over time. The Bone Orientation API is used for calculating bone orientations. Also, it handles the animation by updating the skeleton movements and location with calculated information about transformations, such as translation, rotation and scaling of specific bone, from the relative bone transforms of Kinect skeleton.

Before the 3D model is rendered, it must be positioned in the virtual scene. The relationship between the avatar and the scene is defined and the position and size of the model in the scene is determined. Using the Color Stream, the Kinect video sequence is captured and set as background in the scene. Color video frame is stored and rendered as texture. XNA Skinning processor from Microsoft XNA Framework 4.0 [31] is used for rendering the 3D model skinned mesh on the screen, together with texture, lighting and optional shader effects.

*C. Concealing the de-identified data in videos*

To preserve and conceal the identity of the person in front of the Kinect device, we use the LSB insertion [28] steganographic algorithm. The picture of the virtual character rendered over the color image is used as vessel data or carrier image [32]. The part of the original image containing the person is embedded into the carrier image.

The LSB insertion algorithm [28] implements the idea of hiding one image inside another by using the least significant bits of one image to hold information about the most significant bits of another, secret image. The intuition of the approach is that with a change of the least significant bits, color value changes for each of the three components would be minor, and the change in resulting color would not be conspicuous. If, for example, 4 bits are used for hiding the information, and one of pixel's components, such as blue, has a binary value of 11100001 (decimal 225), first 4 bits will be kept unchanged. Other 4 least significant bits will be replaced with 4 most significant bits from the secret image which will be hidden. If the value of the blue component of corresponding pixel in secret image is 10001101 (decimal 139), the combined picture, so called *stego-media*, will have the value of blue component of specific pixel set as 10111000 (decimal 232). Fig. 4 shows the process of replacing 4 bits of blue component of carrier image, with corresponding bits of secret image.

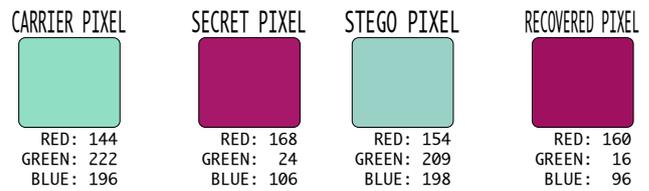

Fig. 5: Concealing the secret image in the carrier image using LSB insertion.

If we examine the numeric values of the resulting color combination of the R, G and B components we can notice that they have not changed much. Fig. 5 shows how pixels of different color can be combined without major changes in carrier pixel. The hidden image can be restored from stego-media using the reversed process. However, the values will be slightly different from the original values. In the process of extracting the hidden image, the last 4 least significant bits of stego image will be used as most significant bits of recovered image. The remaining least significant bits will be filled with zeros. The original image can also be recovered from the stego image. It is done by removing the exact number of least significant bits and filling the gaps with zeros. To process the whole image, it is necessary to loop over the combined image pixels and process all of the 3 components – red, green and blue.

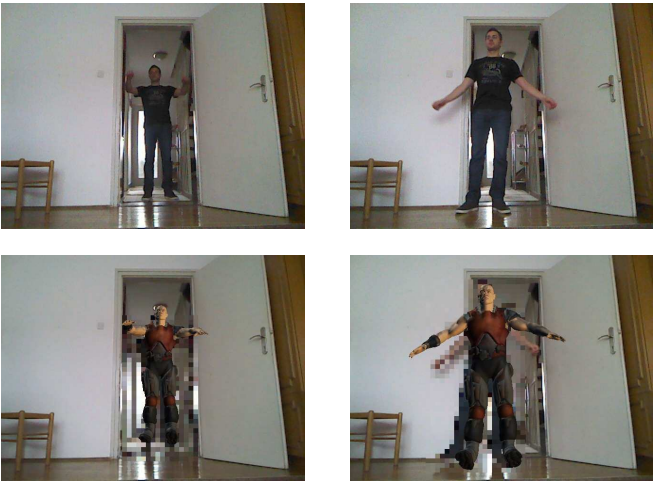

Fig. 6: Top: two sample images captured by the Kinect device. Bottom: rendered avatars and pixelated silhouettes.

## IV. Experiments

In this section, we qualitatively evaluate the output of the two stages of our system: (i) joint detection and 3D model rendering, and (ii) steganographic encoding.

### A. Model rendering

Two example images produced by our system are shown in Fig. 6. As can be seen from Fig. 6, the rendered virtual character does not always completely cover the person in the video. This is due to erroneous output of the Kinect skeletal tracking system, as some joints can be incorrectly localized. Consequently, the positioning of the 3D model in the virtual scene can then be wrong.

In order to improve the results, we introduce the pixelization of the human silhouette, i.e. we pixelize the sections of the color image where the person is visible. By additionally blurring the image, the person is made unrecognizable. The color image itself is retrieved from the color stream which stays intact. For tracking the position of the skeleton another stream is used, and because of that pixelization or any other kind of image editing does not affect the movement of 3D model, so information about the movement and gestures of the person in camera's FOV is preserved.

The exact location of the person in the video frame is determined by using the Coordinate mapping API. Skeleton points are mapped to depth points, and assuming that resolution of depth and color images is identical, the screen coordinates of an object, or in our case the person, can be extracted. In the color image, pixel values are changed in the area that surrounds the person. The image processed this way is then used as background in the virtual scene, so when rendered, the 3D model covers this exact image, as can be seen in Fig. 6.

### B. Steganographic encoding

As described previously, for steganographic encoding we use the LSB insertion algorithm [28]. The original image has a resolution of $640 \times 480$ pixels, with 8 bits per component. For output, we use the uncompressed Bitmap file format, which is suitable for easily storing and extracting the additional data.

Fig. 7 illustrates the quality of the stego image and the recovered image when using a variable number of bits for storing information. By increasing the number of bits for storing the secret image data, the quality of the secret image will increase. By using 5 or more bits of the original image to store the secret image, the stego image quality drops, while at the same time the quality of the recovered image increases. The quality of these two images is leveled when using 4 bits for hiding data. Stego image is then similar to the carrier image and the degradation of quality is not visible. It is also possible to recover a high-quality hidden image.

## V. Conclusion

The solution presented in this paper offers detecting and tracking the movement of a person and hiding their identity. The motion of the person is captured using Kinect. The device sends the information about the skeleton position and movement of the bones which is then mapped to the movement of a virtual 3D humanoid model placed inside a virtual scene. In the scene itself, each frame of the color stream from Kinect is saved as texture and displayed in the background. The 3D model is always positioned closer to the camera, following the position of user, so when it is rendered, it covers the person's identity, but the movement and gestures are preserved.

The aim of this paper was to develop an application that will hide the identity of person by using a digital avatar and accurate tracking of person's movement. The goal was achieved, but some limitations still exist. The 3D model cannot perform a broad range of body movements. Actions and motions like jumping, rotating in position and crouching are not supported. There are some additional limits which pertain to the Kinect device. Two cameras, RGB and IR, are not physically located in the same spot, so a pixel in the image from one camera does not depict the same thing as a pixel in the image from the other camera, even though their screen coordinates are identical. This can cause errors in tracking of movements, which will then be mapped to a motions of a 3D model. Also, sudden moves will not be tracked accurately. Another restriction is the one related to the distance between objects and sensor. If an object or a person is too far away from the sensor, the camera will not recognize it. The same applies to an object being too close.

For the scene to be more natural, we used a humanoid avatar, although we are planning to replace it in the future improvements with an even more realistic one. In future research, we will also consider using multiple Kinect devices for capturing the joints data.

## Acknowledgments

This work has been supported by the Croatian Science Foundation, within the project "De-identification Methods for Soft and Non-Biometric Identifiers" (DeMSI, UIP-11-2013-1544). This support is gratefully acknowledged.

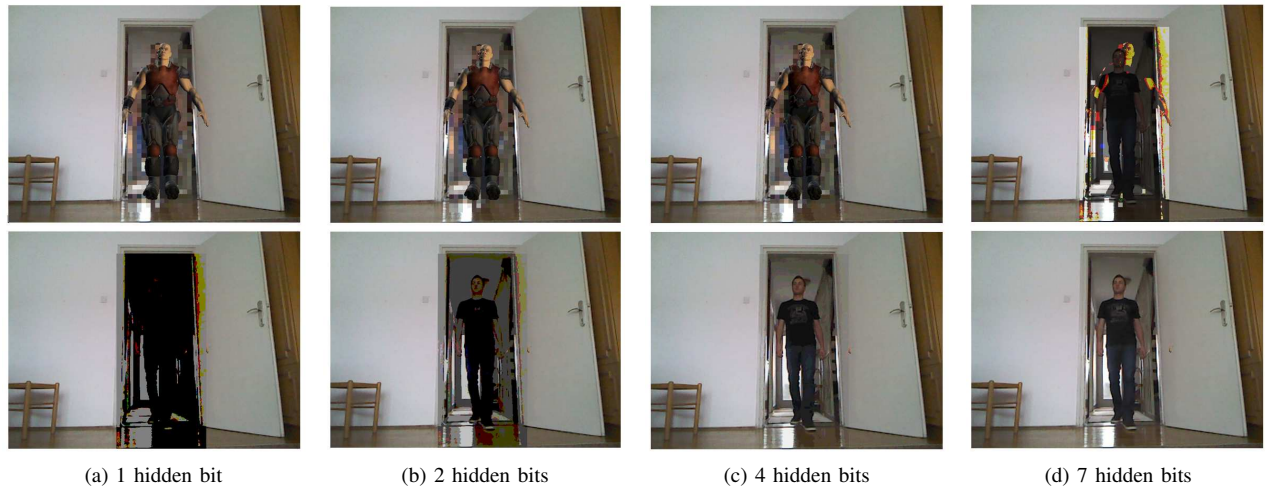

(a) 1 hidden bit   (b) 2 hidden bits   (c) 4 hidden bits   (d) 7 hidden bits

Fig. 7: Pairs of stego images (top) and recovered images (bottom) when using variable number of bits for storing the secret information.